\pdfoutput=1

\documentclass[11pt]{article}

\usepackage[preprint]{acl}

\usepackage{times}
\usepackage{latexsym}

\usepackage[T1]{fontenc}

\usepackage[utf8]{inputenc}
\usepackage{xcolor} 

\usepackage{amsmath,amsfonts,bm}

\def\eqref#1{equation~\ref{#1}}

\def\1{\bm{1}}

\DeclareMathAlphabet{\mathsfit}{\encodingdefault}{\sfdefault}{m}{sl}
\SetMathAlphabet{\mathsfit}{bold}{\encodingdefault}{\sfdefault}{bx}{n}

\definecolor{myblue}{RGB}{0, 0, 255} 
\definecolor{mydarkblue}{RGB}{0, 0, 139} 

\usepackage{url}            
\usepackage{siunitx}
\usepackage{tikz}
\usepackage{wasysym}
\usepackage{booktabs}       
\usepackage{amsfonts}       
\usepackage{nicefrac}       
\usepackage{microtype}      
\usepackage{adjustbox}

\usepackage{graphicx}
\usepackage{tabularx}
\usepackage{subcaption}
\usepackage{multicol}

\usepackage{amsmath}
\usepackage{amssymb}
\usepackage{mathtools}
\usepackage{amsthm}
\usepackage{algorithm}
\usepackage{algorithmic}

\usepackage{multirow}
\usepackage{bm}
\usepackage{colortbl}
\usepackage{wrapfig}
\usepackage{enumitem}
\usepackage{soul}
\usepackage{makecell}
\usepackage{thm-restate}
\usepackage{makecell}   

\usepackage[capitalize,noabbrev,nameinlink]{cleveref}

\usepackage{microtype}

\usepackage{inconsolata}

\usepackage{graphicx}

\makeatletter 
\newcommand{\fixed@sra}{$\vrule height 2\fontdimen22\textfont2 width 0pt\rightarrow$}
\newcommand{\shortarrow}[1]{%
  \mathrel{\text{\rotatebox[origin=c]{\numexpr#1*45}{\fixed@sra}}}
}
\makeatother

\definecolor{darkred}{rgb}{0.80, 0.0, 0.0}
\definecolor{royalblue}{rgb}{0.0549, 0.9118, 0.9224} 
\definecolor{forestgreen}{rgb}{0.1333, 0.5451, 0.1333} 
\definecolor{grn}{rgb}{0.1, 0.6, 0.1}
\definecolor{mgt}{rgb}{0.7, 0.3, 0.7}
\definecolor{chamoisee}{rgb}{0.63, 0.47, 0.35}
\definecolor{purp}{rgb}{0.65, 0.16, 0.65}
\definecolor{alizarin}{rgb}{0.82, 0.1, 0.26}
\definecolor{azure(colorwheel)}{rgb}{0.0, 0.5, 1.0}
\definecolor{brown}{rgb}{0.65, 0.16, 0.16}
\definecolor{lblue}{rgb}{0, 0.2, 0.8}
\definecolor{orange}{rgb}{1.0, 0.5, 0.0}

\definecolor{darkblue-purple}{RGB}{48, 35, 122}
\definecolor{cyan}{RGB}{0, 100, 138}
\definecolor{darkgreen}{rgb}{0.0, 0.5, 0.0}
\definecolor{darkyellow}{rgb}{0.8, 0.6, 0.1}
\definecolor{myblue}{RGB}{0, 0, 255} 
\definecolor{mydarkblue}{RGB}{0, 0, 139} 

\theoremstyle{plain}

\theoremstyle{definition}

\theoremstyle{remark}

\title{Can LLMs Generate Diverse Molecules? \\ Towards Alignment with Structural Diversity}

\author{%
  Hyosoon Jang$^1$, Yunhui Jang$^2$, Jaehyung Kim$^3$, Sungsoo Ahn$^2$ \\
  $^1$POSTECH\quad$^2$KAIST\quad$^3$Yonsei University \\
  \texttt{hsjang1205@postech.ac.kr,} \texttt{jaehyungk@yonsei.ac.kr,} \\ \texttt{\{yunhuijang,sungsoo.ahn\}@kaist.ac.kr} 
}

\begin{document}

\maketitle

\begin{abstract}
Recent advancements in large language models~(LLMs) have demonstrated impressive performance in molecular generation, which offers potential to accelerate drug discovery. However, the current LLMs overlook a critical requirement for drug discovery: proposing a diverse set of molecules. This diversity is essential for improving the chances of finding a viable drug, as it provides alternative molecules that may succeed where others fail in real-world validations. Nevertheless, the LLMs often output structurally similar molecules. While decoding schemes like diverse beam search may enhance textual diversity, this often does not align with molecular structural diversity. In response, we propose a new method for fine-tuning molecular generative LLMs to \textit{autoregressively generate a set of structurally diverse molecules}, where each molecule is generated by conditioning on the previously generated molecules. Our approach consists of two stages: (1) supervised fine-tuning to adapt LLMs to autoregressively generate molecules in a sequence and (2) reinforcement learning to maximize structural diversity within the generated molecules. Our experiments show that the proposed approach enables LLMs to generate diverse molecules better than existing approaches for diverse sequence generation.
\end{abstract}

\section{Introduction}
\label{sec:intro}



Recent advances in large language models (LLMs) have demonstrated the potential to accelerate scientific discovery by leveraging their language processing capabilities. This progress has been particularly impactful for candidate design problems such as drug discovery~\citep{pei2024biot5plus}, protein design~\citep{zhuo2024protllm}, and material design~\citep{gruver2024finetuned}. In particular, with biomolecular datasets and molecular string representations, e.g., SMILES~\citep{weininger1988smiles} or SELFIES~\citep{krenn2020self}, LLMs have demonstrated impressive abilities to generate molecules from textual descriptions, e.g., molecular properties~\citep{edwards2022molt5,ye2023drugassist,pei2024biot5plus}.

However, current LLM-based molecular generation approaches~\citep{edwards2022molt5,ye2023drugassist,pei2024biot5plus} often overlook a critical requirement for drug discovery:\textit{ proposing a diverse set of molecules}. In computer-aided drug discovery, identifying a single molecule with a desired property does not guarantee success in real-world pipelines that require additional cell-based studies and clinical trials~\citep{vamathevan2019applications}. Therefore, drug discovery requires a set of structurally diverse molecules. The generation of structurally diverse molecules increases the chances of finding a viable drug candidate \citep{xie2023how}, as different molecules may succeed where others fail. This diversity is essential to enhance the robustness and success of the drug discovery \citep{krantz1998diversification,hong2020late,sadybekov2023computational}.

\definecolor{OliveGreen}{rgb}{0.56,0.76,0.52}
\definecolor{OrangeRed}{rgb}{0.97,0.58,0.58}
\definecolor{Cornflower}{rgb}{0.66,0.66,0.97}

\begin{figure*}[t]
\begin{subfigure}[t]{\linewidth}
\centering
  \includegraphics[width=0.9\linewidth]{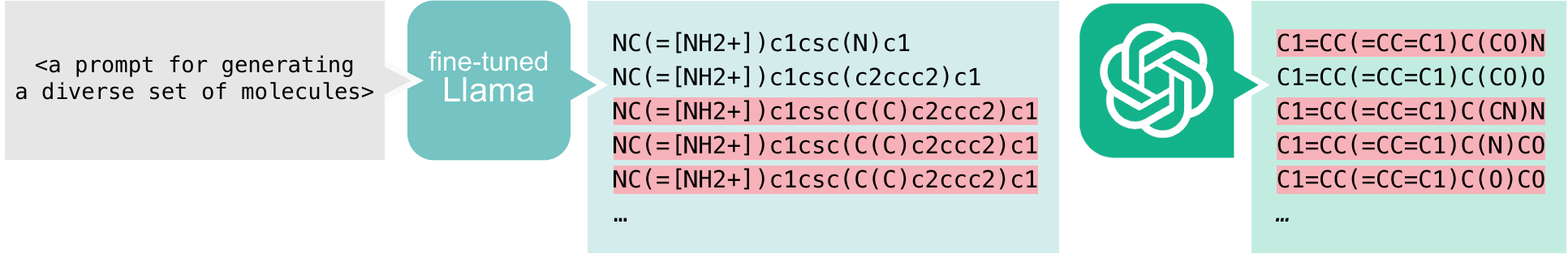} 
  \vspace{-.01in}
  \caption{Existing LLMs \citep{ye2023drugassist,openai2023chatgpt} lack the ability to generate a diverse set of molecules.}\label{subfig:llm_fail}
\end{subfigure}
\centering
\begin{subfigure}[t]{\linewidth}
\centering
\vspace{.15in}
  \includegraphics[width=0.903\linewidth]{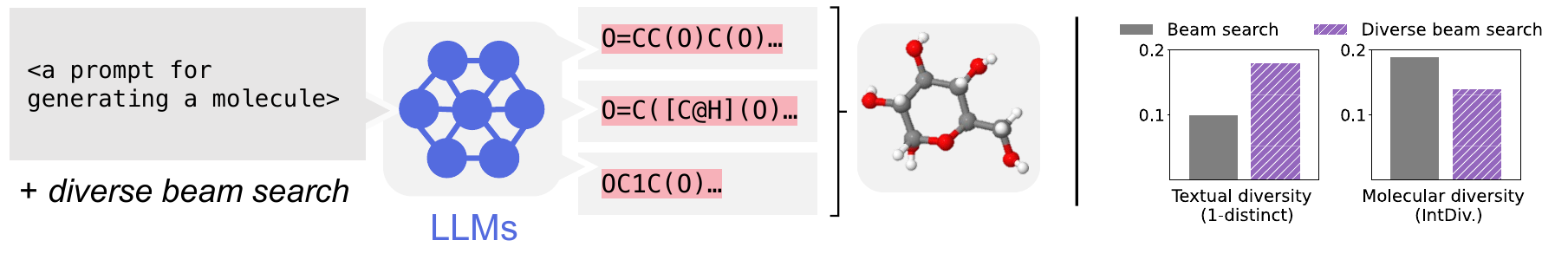} 
  \vspace{-.1in}
  \caption{\textbf{(Left)} Diverse output sequences (SMILES) induce the same molecular structures. \textbf{(Right)} Improved textual diversity via diverse beam search does not enhance molecular diversity in the experiments (\Cref{subsec:chebi-20}).}\label{subfig:div_seq_vs_div_mol}
\end{subfigure}
\caption{\textbf{Existing works on LLMs fail to generate diverse molecules.} The existing decoding schemes \citep{Vijayakumar_Cogswell_Selvaraju_Sun_Lee_Crandall_Batra_2018} for diverse sequence generation and LLMs for chemical tasks fail to capture the molecular diversity, and may induce \sethlcolor{OrangeRed}\hl{structurally identical molecules}.}\label{fig:example}
\vspace{-.15in}
\end{figure*}

In response, we explore the use of LLMs for diverse molecular generation. We begin by identifying the limitations of recent LLMs \citep{ye2023drugassist, openai2023chatgpt} and decoding schemes \citep{Vijayakumar_Cogswell_Selvaraju_Sun_Lee_Crandall_Batra_2018, su2022contrastive} in generating diverse molecules. Then, we present a new method for fine-tuning LLMs to generate diverse molecules. Our approach can be broadly applied to other LLM-based candidate design problems, e.g., computer-aided design \citep{wu2023cadllm}.

\noindent \textbf{Existing LLMs have limitations in generating diverse molecules.} To obtain diverse molecules, one may query the recent generalist LLMs, e.g., Llama~\citep{touvron2023llamaopenefficientfoundation} or ChatGPT \citep{openai2023chatgpt}. However, our empirical observation in \cref{subfig:llm_fail} reveals that even the recent models produce structurally identical or highly similar molecules from the given prompt.\footnote{ChatGPT-4o \citep{openai2023chatgpt} generates different SMILES strings that map to an identical molecule.} This observation aligns with previous observations that have shown LLMs may fail to generate diverse outputs~\citep{kirkunderstanding} for general text-based domains.


\noindent \textbf{Decoding schemes for diversified generation do not align with molecular diversity.} We also acknowledge the existence of decoding schemes, e.g., diverse beam search \citep{Vijayakumar_Cogswell_Selvaraju_Sun_Lee_Crandall_Batra_2018} or contrastive beam search \citep{su2022contrastive}, which have been proposed to improve the diversity of output sequences generated by LLMs. However, these decoding schemes are limited to improving the textual diversity which often does not correspond to molecular structural diversity, e.g., there exist many SMILES or SELFIES strings that correspond to the same molecule, as illustrated in \Cref{subfig:div_seq_vs_div_mol}.



\noindent \textbf{Our approach.} We repurpose existing molecular generative LLMs to autoregressively generate a diverse set of molecules from a single prompt. By enabling the LLMs to generate a new molecule conditioned on previously generated molecules, we expect the LLMs to learn to enhance the structural diversity between the generated molecules. To this end, we propose a two-stage approach to fine-tune LLMs: (a) a supervised fine-tuning stage to repurpose LLMs to autoregressively generate a sequence of multiple molecules and (b) a reinforcement learning stage to maximize the molecular structural diversity between the generated molecules. 


In the supervised training stage, we train LLMs to autoregressively generate a set of molecules in a sequence. Note that the training data, i.e., a set of molecules, can be collected from LLMs themselves through iterative sampling, and then filtered to enhance the quality, e.g., removing invalid molecules. However, this stage does not necessarily incorporate molecular diversity, as the training may not involve sufficiently distinct molecules (e.g., limitations in \Cref{subfig:div_seq_vs_div_mol}). To tackle this, we subsequently apply reinforcement learning with exploration towards discovering diverse molecules.

Next, in the reinforcement learning stage, we train LLMs to maximize the diversity of molecules within a generated sequence. However, for our task, conventional sequence-wise reinforcement learning~\citep{ouyang2022training} suffers from the credit assignment problem \citep{zhou2024archer}: the challenges in identifying and promoting the generation of molecules responsible for increasing diversity, among a larger set of molecules in the sequence. To resolve this issue, we solve multi-stage molecule generation problems for a sequence of molecules, where the generation of each molecule aims to maximize the diversity with respect to the previously generated molecules. We train LLMs to maximize the associated rewards using proximal policy optimization \citep{schulman2017proximal}.



We compare our method with the decoding schemes for diversified generation \citep{vijayakumar2016diverse,su2022contrastive} and other representative LLMs, including chemical-task specialists \citep{edwards2022molt5,christofidellis2023unifying,pei2023biot5,pei2024biot5plus}, fine-tuned generalists on chemical domains \citep{fang2023mol-inst,yu2024llasmol}, and the ChatGPT series \citep{openai2023chatgpt,openai2024o1}. We observe that (1) our fine-tuning approach enables LLMs to better discover diverse molecules compared to existing decoding schemes and (2) our fine-tuned LLM outperforms other LLMs.

To conclude, our contributions can be summarized as follows:
\begin{itemize}[topsep=-1.0pt,itemsep=1.0pt,leftmargin=3.5mm]
\item We are the first to explore the use of LLMs for generating diverse molecules. 
\item We first propose a fine-tuning approach for LLMs to generate diverse solutions, which presents a new direction distinct from existing approaches focused on the decoding scheme.
\item Experimentally, our method outperforms the baselines in generating diverse molecules.
\end{itemize}

\section{Related Work}
\label{sec:related}

\noindent \textbf{Large language models (LLMs) for molecular generation.} Recent advancements in LLMs have shown increasing promise in scientific applications, especially for molecular generation \citep{edwards2022molt5,pei2023biot5,fang2023mol-inst,pei2024biot5plus}. First, \citet{edwards2022molt5} proposed MolT5, a molecular generative LLM that translates between SMILES \citep{weininger1988smiles} and molecular text descriptions. Next, Text+Chem T5 \citep{christofidellis2023unifying} and BioT5 \citep{pei2023biot5} considered pre-training molecular generative LLMs on datasets that incorporate extensive chemical knowledge, e.g., scientific articles. Additionally, \citet{ye2023drugassist}, \citet{fang2023mol-inst}, \citep{chen2023meditron}, and \citet{yu2024llasmol} fine-tuned generalist LLMs, e.g., Llama \citep{touvron2023llamaopenefficientfoundation}, through biological instructions, molecular modifications, and large-scale molecular datasets, respectively.

\noindent \textbf{Decoding schemes for generating diverse output sequences.} To generate diverse solution candidates from LLMs, existing literature on LLMs has studied improving decoding schemes. To acquire multiple distinct sequences with high likelihoods, one can consider employing beam search, which jointly decodes multiple distinct outputs \citep{och-2003-minimum}. To enhance diversity between sequences, \citet{vijayakumar2016diverse,Vijayakumar_Cogswell_Selvaraju_Sun_Lee_Crandall_Batra_2018} incorporated token-wise differences between generated sequences in the beam search. Furthermore, \citet{su2022contrastive} considered the contrast between the candidate sequences. In addition, \citet{Holtzman2020The} proposed nucleus sampling, which enhances random sampling by balancing the quality and the diversity.

\noindent \textbf{Reinforcement learning (RL) for fine-tuning LLMs.} RL has been effectively applied to fine-tune LLMs, aligning them with desired behaviors expressed through reward signals. One notable example is RL from human feedback to align LLMs with human preference \citep{ouyang2022training}. In addition, there has been a surge in research on devising RL for LLMs as well, such as addressing multi-turn settings \citep{shani2024multi} and incorporating multiple fine-grained reward signals \citep{wu2023finegrained}. For molecular generation, \citet{ghugare2024searching} proposed RL-based fine-tuning to generate a molecule satisfying target properties. However, to the best of our knowledge, there exist no prior RL-based approaches that aim to increase the diversity of LLM-generated outputs.\footnote{{More related works, e.g., RL for diverse molecular generation without using LLMs, are described in \Cref{appx:addition_related}.}}

\begin{figure*}[t]
\centering
\begin{subfigure}[t]{\linewidth}
\centering
  \includegraphics[width=0.9\linewidth]{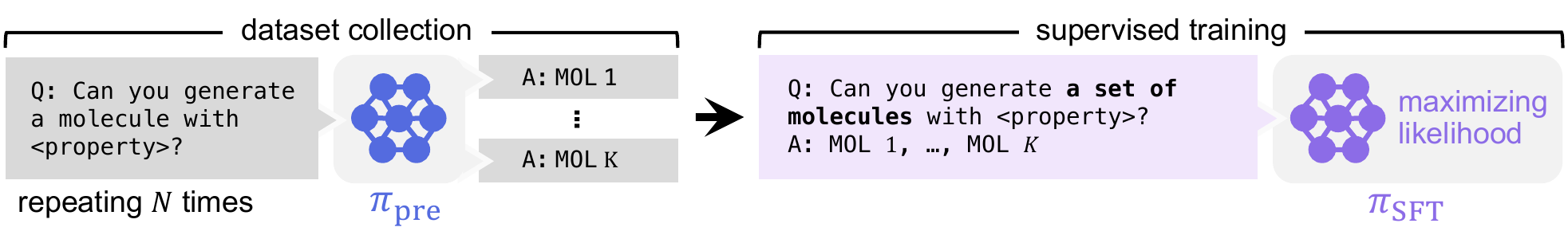}
  \vspace{-.02in}
  \subcaption{\textbf{Supervised fine-tuning (\Cref{alg:algorithm}):} enables generating a sequence of multiple molecules.}\label{subfig:SFT}
\end{subfigure}
\begin{subfigure}[t]{\linewidth}
\centering
  \includegraphics[width=0.9\linewidth]{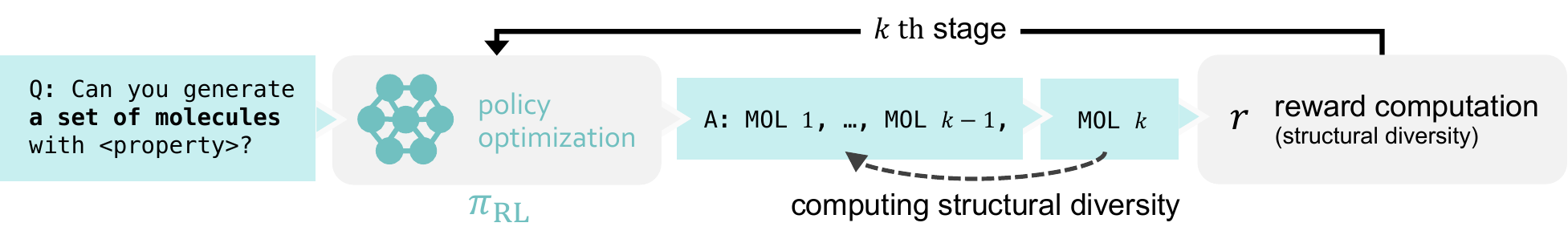}
  \vspace{-.05in}
 \subcaption{\textbf{Reinforcement learning (\Cref{alg:rl_algorithm}):} maximizes the diversity between new and previously generated molecules.}\label{subfig:RL}
\end{subfigure}
\caption{\textbf{Illustration of proposed fine-tuning approaches.} We consider two stages for fine-tuning LLMs: a supervised fine-tuning \Cref{subfig:SFT} and a reinforcement learning \Cref{subfig:RL}. The prompts are simplified for explanatory purposes, and the actual prompts are provided in \Cref{appx:decoding}.}\label{fig:overall}
\end{figure*}

\section{Method}

In this section, we present our method for fine-tuning LLMs to generate diverse molecules. Specifically, we consider fine-tuning existing molecular generative LLMs that produce molecular representations such as SMILES or SELFIES. Importantly, our approach is versatile and can be applied to other domains, e.g., protein sequence \citep{zhuo2024protllm} or computer-aided design \citep{wu2023cadllm}. 

\noindent \textbf{Overview.} Our goal is to generate a sequence of structurally diverse molecules from a given prompt by producing them in a single concatenated output. To this end, we fine-tune the LLMs in two stages: (a) a supervised fine-tuning phase that repurposes the LLMs to generate a sequence of molecules rather than a single one, and (b) a reinforcement learning phase aimed at further enhancing the structural diversity among the generated molecules.

\noindent \textbf{Task details.} In detail, we consider generating molecules from a prompt $p_{\text{desc}}$, where the prompt describes a molecular property that the generated molecules should possess. In this setting, we aim to generate diverse molecules that satisfy the given description $p_{\text{desc}}$, where the diversity is evaluated using similarity measures between the structural features of the molecules, e.g., the presence of specific atoms, or substructures \citep{bajusz2015tanimoto}. We let $\mathcal{P}$ denote the prompts used for training.

\subsection{Supervised Fine-tuning} 


\begin{algorithm}[h]
\caption{Supervised fine-tuning}\label{alg:algorithm}
\begin{algorithmic}[1]
   \STATE Initialize $\pi_{\text{SFT}}$ with $\pi_{\text{pre}}$
   \REPEAT 
   \STATE Get prompt $p_{\text{desc}} \sim \mathcal{P}$ 
   \STATE Get $\{m_i\}_{i=1}^{T}$ from $\pi_{\text{pre}}(m \; | \;p_{\text{desc}})$
   \STATE Update $\{m_i\}^{K}_{i=1}\leftarrow \text{Filter}(\{m_i\}^{T}_{i=1})$
   \STATE Maximize \Cref{eq:log} with $\{m_i\}^{K}_{i=1}$
   \UNTIL{Converged}
   \STATE \textbf{Output:} fine-tuned $\pi_{\text{SFT}}$
\end{algorithmic}
\end{algorithm}

\noindent We first describe our supervised fine-tuning process for repurposing the pre-trained LLMs to autoregressively generate multiple molecules in a sequence. This involves collecting a dataset of molecules from a pre-trained LLM $\pi_{\text{pre}}$, and then fine-tuning the LLM $\pi_{\text{SFT}}$ on the collected dataset. We describe the process in \Cref{subfig:SFT} and \Cref{alg:algorithm}.


\noindent \textbf{Dataset collection.} The supervised training process is conducted with a set of training prompts $\mathcal{P}$. Initially, the pre-trained LLM $\pi_{\text{pre}}$ produces a set of molecules by iterative sampling molecules for a given prompt $p_{\text{desc}}\in\mathcal{P}$ as follows:
\begin{equation*}
m_i \sim \pi_{\text{pre}}\left( m_i \middle| p_{\text{desc}} \right)\quad \text{for}\;i=1,\ldots,T,
\end{equation*}
where $m_i$ denotes the string representation of the molecule. In practice, we employ beam search to collect the set of molecules $\{m_i\}^T_{i=1}$. Then, we filter out the invalid string representations, duplicate molecules, and molecules that do not satisfy the given prompt $p_{\text{desc}}$. This results in reducing the set of molecules from $\{m_i\}^T_{i=1}$ to $\{m_i\}^K_{i=1}$. The details are described in \Cref{appx:implementation}.

\noindent \textbf{Supervised training.} Given the filtered set of molecules $\{m_i\}^K_{i=1}$, we train the LLM $\pi_{\text{SFT}}$, which is initialized with $\pi_{\text{pre}}$, to generate them as a single concatenated sequence. We denote this sequence by $\mathcal{M}_{1:K}=m_1||\cdots||m_K$, where $||$ denotes the concatenation of the molecular string representations. Specifically, given a modified prompt $p_{\text{desc+div}}$ describing the target property with a request for generating diverse molecules, we train the LLM to maximize the log-likelihood: 
\begin{equation} \label{eq:log}
\log \pi_{\text{SFT}}\left(\mathcal{M}_{1:K} \; \middle| \; p_{\text{desc+div}}\right). 
\end{equation} 
However, the policy $\pi_{\text{SFT}}$ does not necessarily incorporate a molecular structural diversity, as the set of molecules $\{m_i\}^K_{i=1}$ collected from $\pi_{\text{pre}}$ may insufficiently involve diverse molecular structures~(e.g., due to limitations in \Cref{subfig:div_seq_vs_div_mol}). To tackle this, we next introduce an online reinforcement learning stage with exploration towards discovering diverse molecules.


\subsection{Reinforcement Learning}\label{subsec:RL}

\begin{algorithm}[H]
\caption{Multi-stage RL fine-tuning}\label{alg:rl_algorithm}
\begin{algorithmic}[1]
   \STATE Sample $p_{\text{desc}} \sim \mathcal{P}$ 
    \STATE Sample $m_1 \sim \pi_{\text{RL}}(m_1 \mid p_{\text{desc+div}})$ 
    \STATE Update $\pi_{\text{RL}}$ with PPO to maximize $r(m_1)$
   \FOR{$k=2,\ldots,K$}
        \STATE Get $m_k \sim \pi_{\text{RL}}(m_k \mid \mathcal{M}_{1:k-1}, p_{\text{desc+div}})$ 
       \STATE Update $\pi_{\text{RL}}$ with PPO to maximize $r(m_k)$
       \ENDFOR
\end{algorithmic}
\end{algorithm}

\begin{table*}[t]
\caption{\textbf{Comparison with existing decoding schemes.} \textbf{NCircles} represents both quality and diversity-related metric. \textbf{Accepted \& unique} represent quality-related metrics. \textbf{IntDiv.} represents an average of pair-wise diversities. Our method \textbf{(Div-SFT+RL)} generates more diverse and high-quality molecules compared to the baselines. Notably, our method makes a larger gap over the baselines on \textbf{NCircles} related to capturing both quality and diversity.}\label{tab:biot5_main}
\vspace{-.05in}
\centering
{
\scalebox{0.85}{\begin{tabular}{clcccc}
\toprule
Dataset & Method & NCircles$_{h=0.85}$ & NCircles$_{h=0.75}$ & Accepted \& Unique & IntDiv. \\
\midrule
& Random & 1.079 & 0.948 & 1.970 & 0.109 \\
& Nucleus & 1.006 & 0.918 & 1.623 & 0.090 \\
& BS & 4.562 & 2.603 & 33.406 & 0.176 \\
L+M-24 & Diverse BS & 2.395 & 1.743 & 9.915 & 0.234 \\
& Contrastive BS & 4.521 & 2.594 & \textbf{33.568} & 0.176 \\
& Div-SFT & 5.198 & 3.205 & 20.711 & 0.250 \\
& \textbf{Div-SFT+RL} & \textbf{10.51} & \textbf{6.278} & {31.060} & \textbf{0.287} \\
\midrule
& Random & 1.539 & 1.383 & 1.998 & 0.045 \\
& Nucleus & 0.897 & 0.876 & 0.972 & 0.014 \\
& BS & 5.576 & 4.171 & 11.734 & 0.194 \\
ChEBI-20 & Diverse BS & 4.413 & 3.637 & 5.905 & 0.140 \\
& Contrastive BS & 7.955 & 5.883 & 15.215 & 0.233 \\
& Div-SFT & 4.869 & 3.601 & 8.178 & 0.141 \\
& \textbf{Div-SFT+RL} & \textbf{11.301} & \textbf{8.996} & \textbf{16.271} & \textbf{0.246} \\
\bottomrule
\end{tabular}}
}
\vspace{-.05in}
\end{table*}

\noindent We apply reinforcement learning to maximize the diversity of the generated molecules within a sequence. However, when applied to a sequence of molecules $\mathcal{M}_{1:K}$, conventional sequence-wise reinforcement learning \citep{ouyang2022training} suffers from the credit assignment problem \citep{zhou2024archer}: the challenge in identifying and promoting the generation of molecules responsible for increasing diversity. To circumvent this, we introduce a molecule-wise reinforcement learning.\footnote{We make comparisons between the sequential- and the molecule-wise approaches in \Cref{tab:filtering} of \Cref{subsec:abla}.}

Specifically, we consider reinforcement learning on a sequence of molecules $\mathcal{M}_{1:K}$ as learning in $K$ individual stages. Each stage corresponds to generating a molecule $m_k$ conditioned on a sequence of previously generated molecules $\mathcal{M}_{1:k-1}$. Then, the LLM $\pi_{\text{RL}}$ is trained to maximize the return of each stage, defined by the reward of the generated molecule. Here, the reward is the diversity between the previously generated molecules $\{m_i\}_{i=1}^{k-1}$ and the new molecule $m_k$. We also incorporate an auxiliary reward to ensure that the generated molecule satisfies the description $p_{\text{desc}}$. We present our approach in \Cref{subfig:RL,alg:rl_algorithm}.




\noindent \textbf{Reward.} The reward evaluates the molecule $m_k$ with a diversity reward $r_{\text{div}}(m_k, \{m_i\}^{k-1}_{i=1})$ and a description-matching reward $r_{\text{match}}(m_k,p_{\text{desc}})$:
\begin{equation*} 
r(m_k) = r_{\text{div}}(m_k, \{m_i\}^{k-1}_{i=1}) + r_{\text{match}}(m_k, p_{\text{desc}}), 
\end{equation*} 
where the diversity reward $r_{\text{div}}$ evaluates structural differences between the molecule $m_k$ and the previously generated molecules $\{m_i\}^{k-1}_{i=1}$ by assessing their true molecular structures. Note that $r_{\text{div}}(m_1)$ is zero. The description-matching reward $r_{\text{match}}$ evaluates whether the molecule $m_k$ satisfies the description $p_{\text{desc}}$. In practice, the final action completing the molecule yields the reward. The implementation is described in \Cref{sec:experiment}.

\noindent \textbf{Policy optimization.} We optimize the LLM $\pi_{\text{RL}}$ to maximize the reward using proximal policy optimization \citep[PPO;][]{schulman2017proximal}. Here, $\pi_{\text{RL}}$ is initialized with $\pi_{\text{SFT}}$. We also combine per-token KL penalty from the supervised fine-tuned model following prior studies \citep{ouyang2022training}.

\definecolor{OliveGreen}{rgb}{0.56,0.76,0.52}
\definecolor{OrangeRed}{rgb}{0.97,0.58,0.58}
\definecolor{Cornflower}{rgb}{0.66,0.66,0.97}

\begin{table*}[t]
\caption{\textbf{Visualization of the generated molecules.} We generate ten molecules from: ``The molecule is a egfr inhibitor and modulate, belonging to the cancer treatment class of molecules, and is treatment of disorder''. The \textcolor{blue}{blue line} indicates the molecule that follows the given description, as evaluated using the external tool \citep{trott2010autodock}. BS generates \sethlcolor{OrangeRed}\hl{structurally similar molecules}. The details are described in \Cref{appx:qualit}.}\label{tab:ex_biot5}
\vspace{-.05in}
\centering
{
\scalebox{0.92}{\begin{tabular}{ccc}
\toprule
BS & & Ours \\
\midrule
\includegraphics[width=0.47\linewidth]{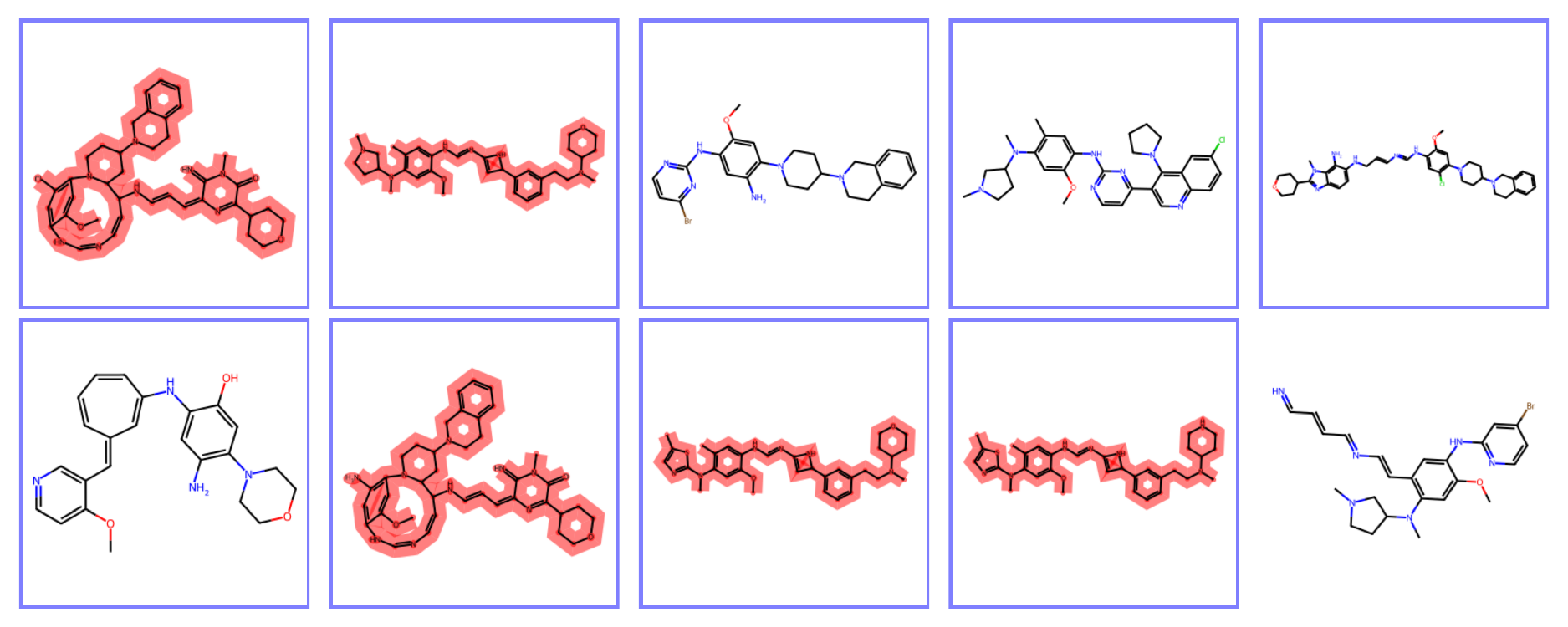} & \;\; & \includegraphics[width=0.47\linewidth]{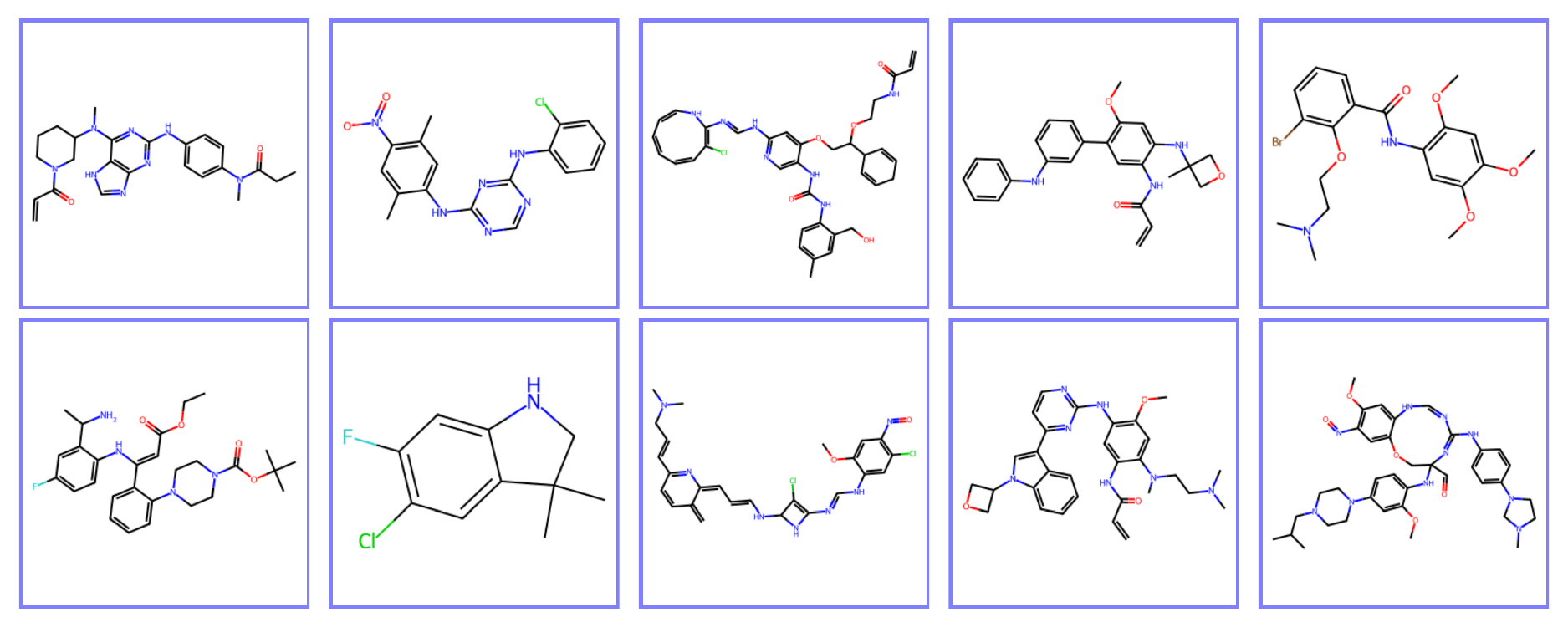} \vspace{-.03in} \\
\bottomrule
\vspace{-.1in}
\end{tabular}
}}\vspace{-.1in}
\end{table*}

\section{Experiment}
\label{sec:experiment}

In this section, we validate our supervised fine-tuning and reinforcement learning methods for diverse molecular generation, coined Div-SFT and Div-SFT+RL, respectively. 

\subsection{Comparison with Decoding Schemes} \label{subsec:chebi-20}

We first show that our fine-tuning method enables molecular generative LLMs to generate diverse molecules better than applying decoding schemes for diverse sequence generation. Here, we implement our method and decoding schemes on a recent description-guided molecular generative LLM: BioT5$^{+}$ \citep{pei2024biot5plus}.

\noindent \textbf{Tasks.} We consider diverse molecular generation with two existing datasets: L+M-24 \citep{edwards-etal-2024-l} and ChEBI-20 \citep{edwards-etal-2021-text2mol}. Each data point in these datasets provides a qualitative molecular property description, e.g., ``The molecule is an EGFR inhibitor'', and some examples of target molecules that satisfy the description.\footnote{In L+M-24 training split, an average of $70$ provided molecules corresponds to each molecular property description. We provide the data statistics in \Cref{appx:stat}.} Both L+M-24 and ChEBI-20 datasets consist of training and test splits where the training splits have been used to pre-train the base LLM. Our fine-tuning method also uses their training splits. We generate $50$ molecules for each description in the test splits for evaluation.\footnote{In \Cref{subsec:abla}, we further validate our method by generating $100$, $150$, and $200$ molecules for each description.}

\noindent \textbf{Metrics.} We evaluate the structural diversity of the generated molecules that satisfy the given molecular description. However, evaluating whether the generated molecule satisfies some qualitative properties, e.g., a membrane stabilizer, is intractable due to the complexity of the required assessments. Therefore, we assume that the molecule satisfies the description if it shares a certain degree of structures with one of the target molecules provided in the dataset.\footnote{In \Cref{tab:ex_biot5,subsec:genral_tune}, we also evaluate whether the generated molecules actually satisfy the molecular description using an external tool, e.g., a docking tool.} Here, the specific degree is described in the explanation of the accepted molecules.

To assess the similarity between two molecules, we compute Dice and Tanimoto similarities on their structural features \citep{bajusz2015tanimoto}, focusing on the degree of shared substructures and overall structural similarity, respectively. See \Cref{appx:diversity} for details. Based on these, we evaluate the following metrics for the generated molecules:
\begin{itemize}[topsep=-1.0pt,itemsep=1.0pt,leftmargin=3.5mm]
\item The number of accepted and unique molecules (Accepted \& Unique) counts unique molecules that share substructures with one of the provided examples of target molecules (as much as Dice similarity higher than $0.7$).
\item The number of circles \citep[NCircles;][]{xie2023how} considers both quality and diversity. Given the set of accepted molecules, NCircles$_{h}$ computes the size of the largest subset of molecules in which no two molecules are structurally similar to each other (as much as Tanimoto similarity below a threshold $h$). 
\item Internal diversity \citep[IntDiv.;][]{moses} measures the average pair-wise structural distances, based on the complement of Tanimoto similarity, between the accepted molecules.

\end{itemize}

\noindent \textbf{Implementations.} For supervised fine-tuning, we collect molecules for each description $p_{\text{desc}}$ in the training splits (a maximum of $100$ molecules per description). Then, the collected molecules are filtered to remove invalid, duplicated, and unaccepted molecules. The description-matching reward $r_{\text{match}}(m_k)$ is defined as the maximum Dice similarity between the molecule $m_k$ and the provided target molecules. Next, the diversity reward $r_{\text{div}}(m_k,\{m_i\}^{k-1}_{i=1})$ is defined as the complement of the maximum Tanimoto similarity between the molecule $m_k$ and the previously generated molecules $\{m_i\}^{k-1}_{i=1}$. We describe detailed implementations in \Cref{appx:implementation}.


\noindent \textbf{Baselines.} We compare our method with various decoding schemes, including random sampling, nucleus sampling \citep{Holtzman2020The}, and beam search \citep[BS;][]{beamsearch1965}. We also consider the variants of BS that promote sequence-level diversity: diverse BS \citep{Vijayakumar_Cogswell_Selvaraju_Sun_Lee_Crandall_Batra_2018} and contrastive BS \citep{su2022contrastive}. 


\noindent \textbf{Results.} In \Cref{tab:biot5_main}, we present experimental results on L+M-24 and ChEBI-20 datasets. One can see that applying our fine-tuning approach shows superior performance in discovering diverse accepted molecules, i.e., yields the highest NCircles, compared to applying existing decoding schemes on the molecular generative LLM. We also present qualitative results in \Cref{tab:ex_biot5}, which evaluate whether the generated molecules actually satisfy the given description using an external tool \citep{trott2010autodock}. One can see that both BS and our method generate molecules that satisfy the given description, but our method generates more diverse molecules compared to BS.

\definecolor{whitegray}{RGB}{243, 243, 243} 

\begin{table*}[t]
\caption{\textbf{Comparison with results obtained by applying existing decoding schemes to each LLM.} The base LLM of our method is BioT5$^+$. For each baseline, we report the best result obtained by one of the existing decoding schemes. Our fine-tuned model shows superiority in generating diverse molecules by yielding highest NCircles.}\label{tab:sota}
\vspace{-.05in}
\centering
{
\scalebox{0.85}{\begin{tabular}{clcccc}
\toprule
Dataset & Method & NCircles$_{h=0.85}$ & NCircles$_{h=0.75}$ & Accepted \& Unique & IntDiv. \\
\midrule
& MolT5 \small{(Diverse BS)} &  {{6.295}} & {{4.308}} & {{25.841}} & {{0.278}} \\
\multirow{2}{*}{L+M-24} & BioT5$^{+}$ \small{(Contrastive BS)} & {{4.562}} & {{2.603}} & {\textbf{33.568}} & {{0.176}} \\
& Meditron \small{(Diverse BS)} & {{4.693}} & {{3.037}} & {{27.837}} & {{0.263}} \\
& \textbf{Ours} & {\textbf{10.511}} & {\textbf{6.278}} & {{31.062}} & {\textbf{0.287}} \\
\midrule
& MolT5 \small{(Diverse BS)}  & {1.902} & {1.509} & {2.798} & {0.101} \\ 
& Text+Chem T5 \small{(Contrastive BS)} & {6.674} & {4.408} & {15.828} & {0.205} \\ 
{ChEBI-20}& BioT5$^{+}$ \small{(Contrastive BS)} & {7.955} & {5.883} & {15.215} & {0.233} \\
& LlaSMol \small{(BS)} & {7.371} & {5.492} & \textbf{16.978} & {0.217} \\
& \textbf{Ours} & {\textbf{11.301}} & {\textbf{8.966}} & {{16.271}} & {\textbf{0.246}} \\
\bottomrule
\end{tabular}}
}
\end{table*}

\begin{figure*}[t]
\centering
\begin{subfigure}[t]{0.24\linewidth}
\centering
  \includegraphics[width=0.95\linewidth]{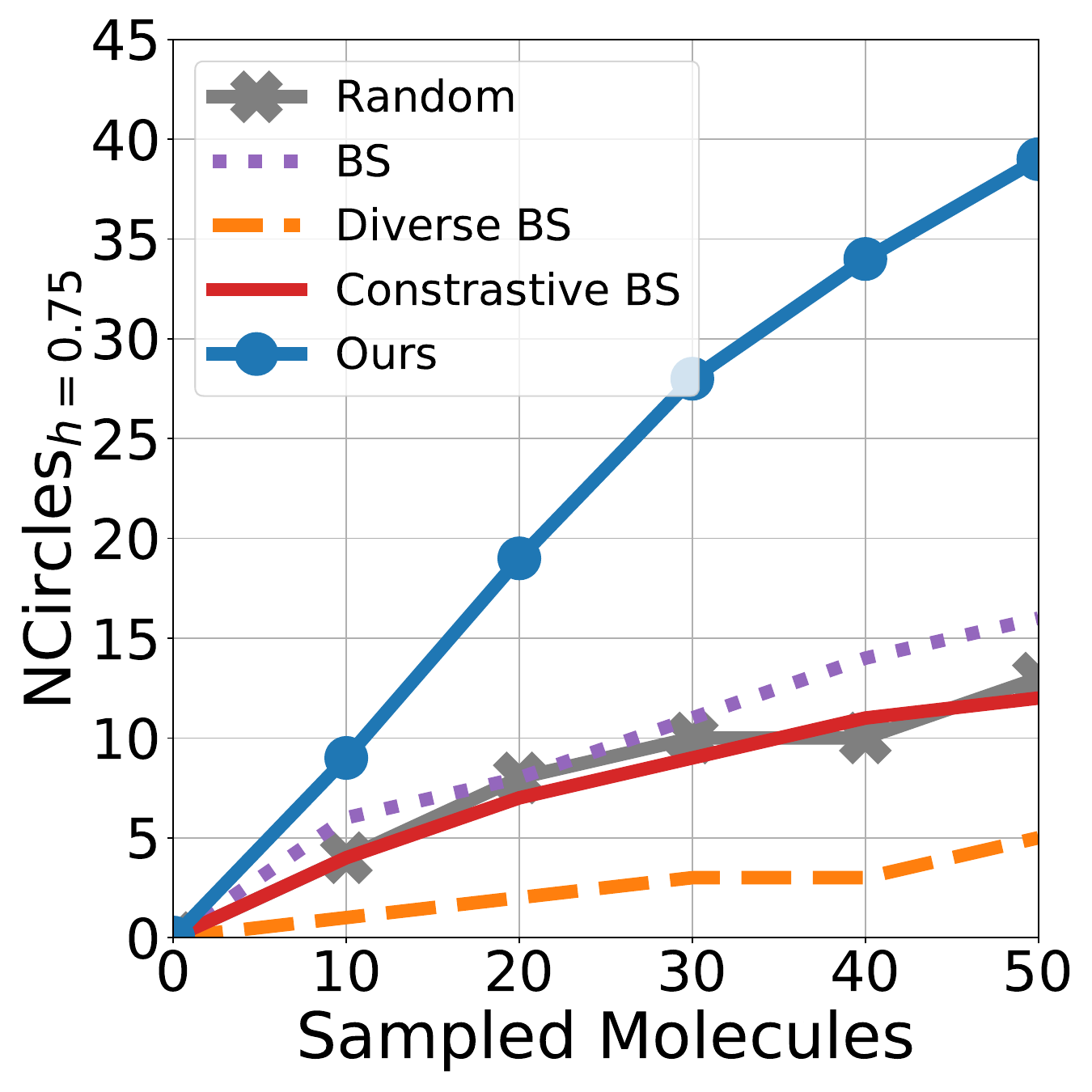}
\vspace{-.05in}
  \subcaption{HB donors}
\end{subfigure}
\begin{subfigure}[t]{0.24\linewidth}
\centering
  \includegraphics[width=0.95\linewidth]{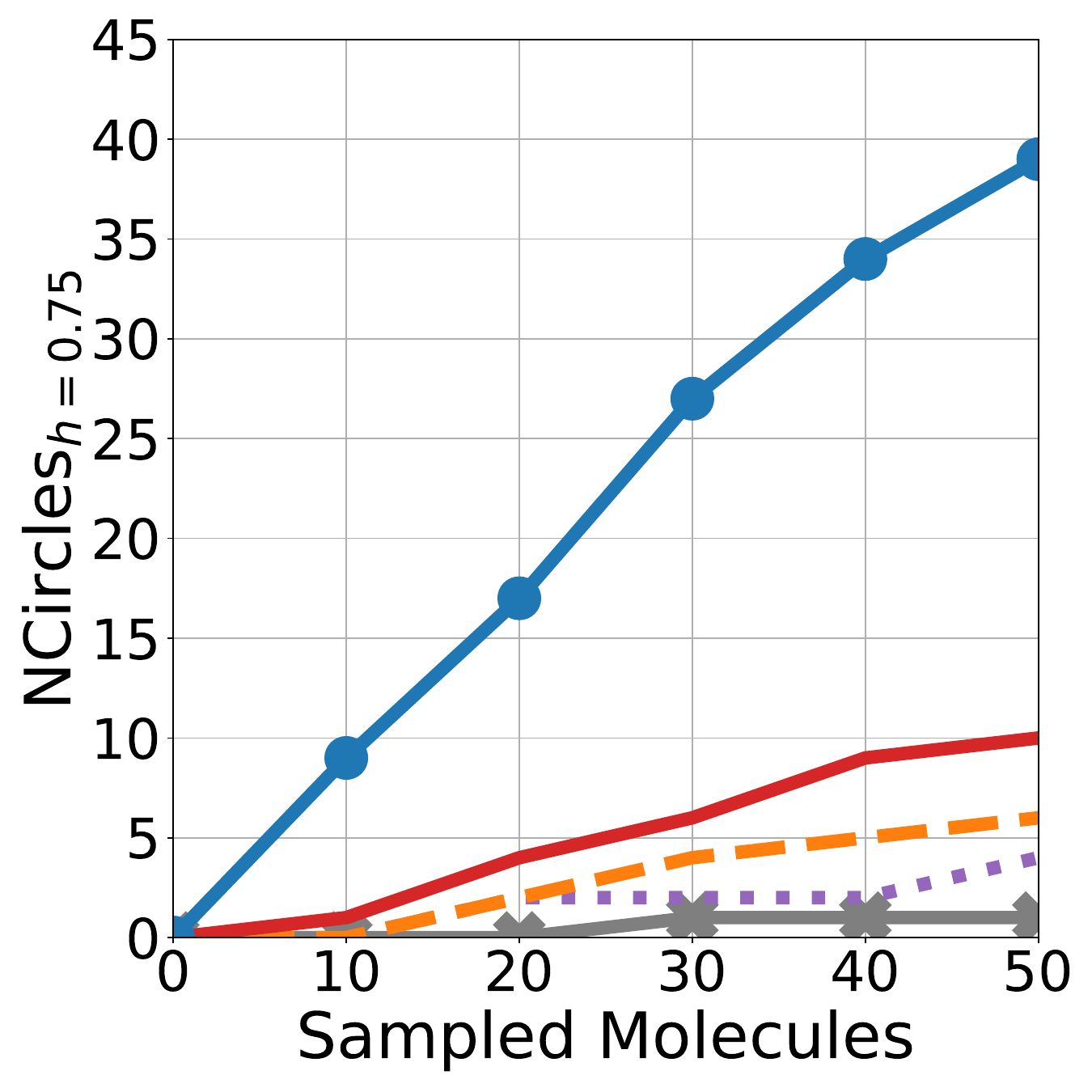}
\vspace{-.05in}
  \subcaption{HB acceptors}
\end{subfigure}
\begin{subfigure}[t]{0.24\linewidth}
\centering
  \includegraphics[width=0.95\linewidth]{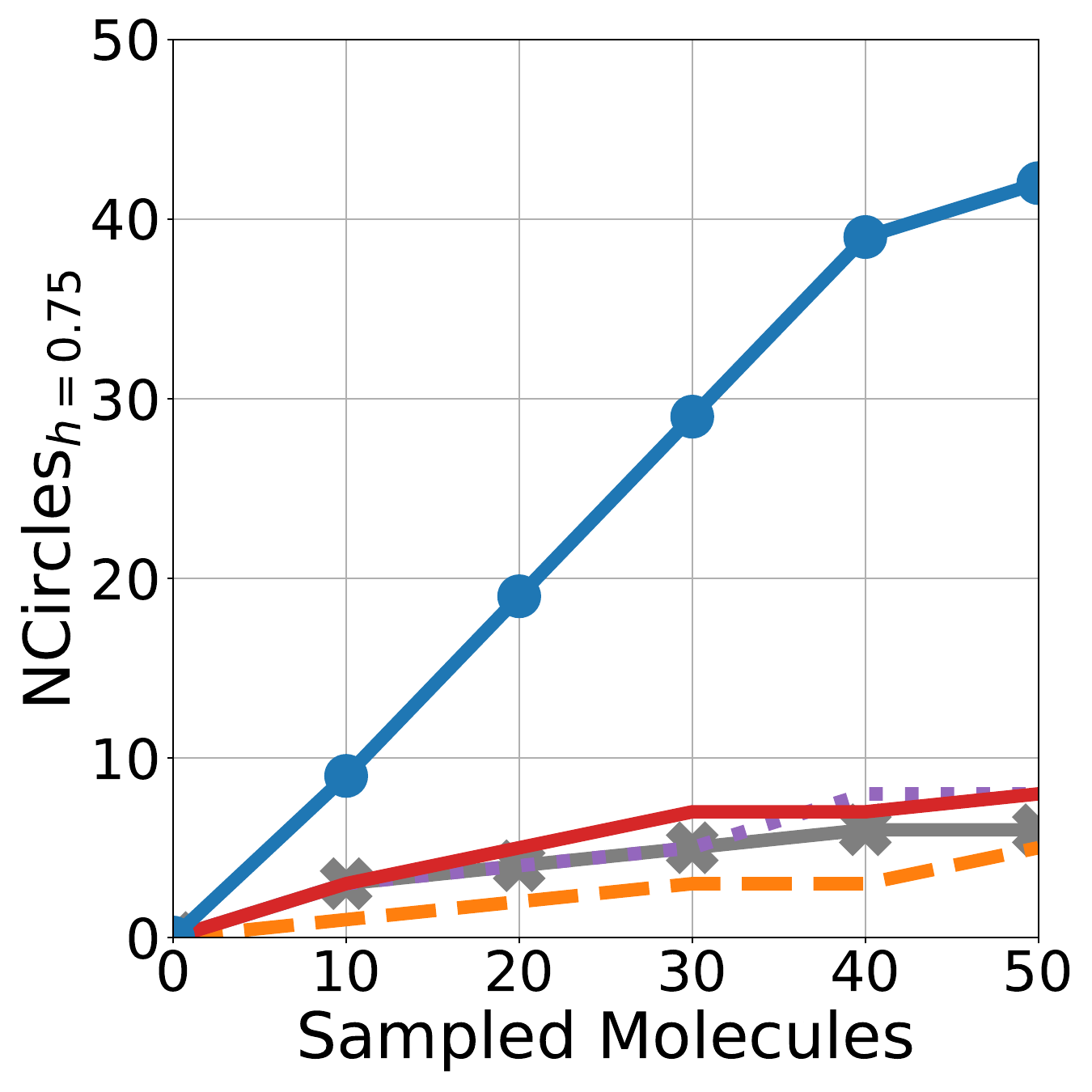}
\vspace{-.05in}
  \subcaption{Bertz complexity}
\end{subfigure}
\begin{subfigure}[t]{0.24\linewidth}
\centering
  \includegraphics[width=0.95\linewidth]{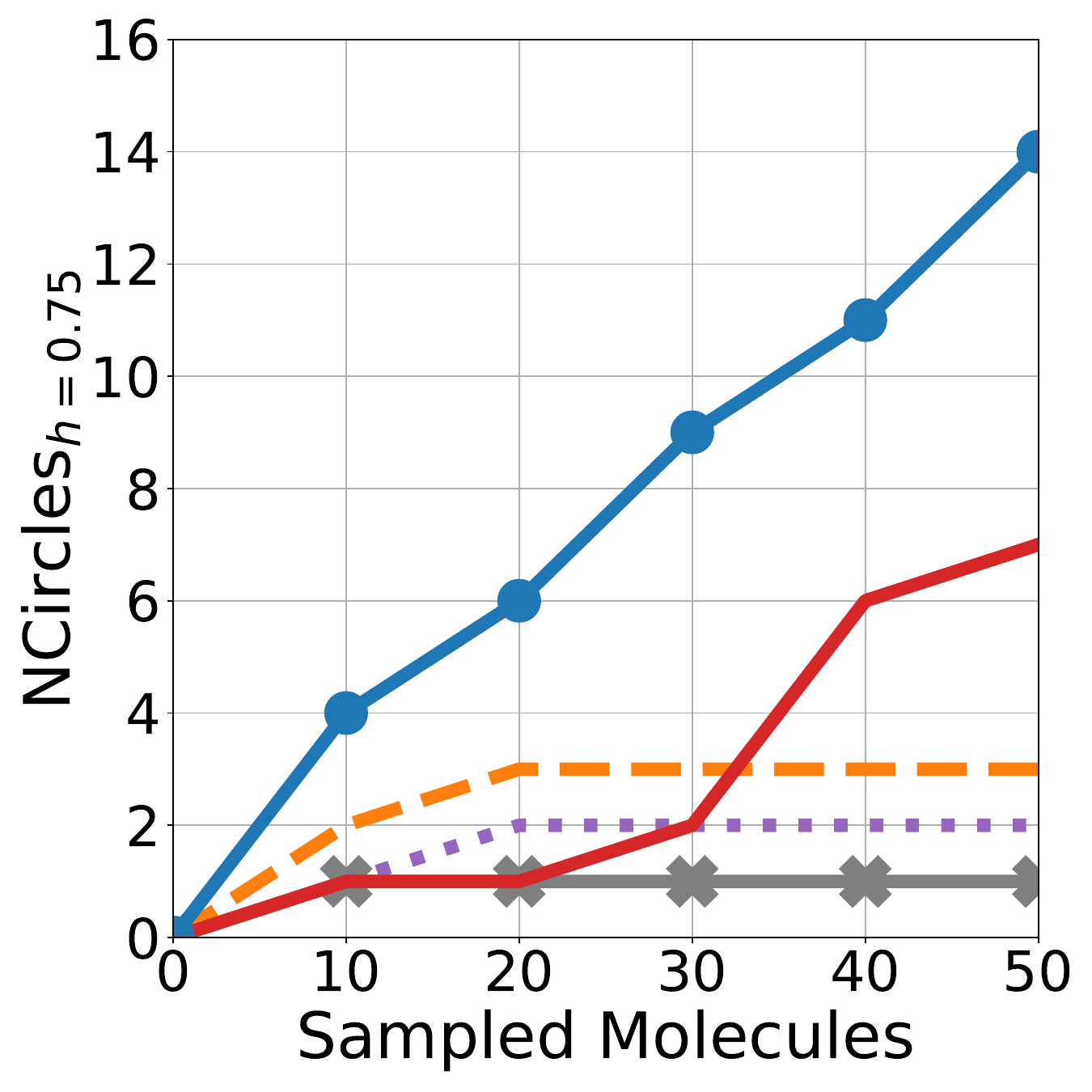}
\vspace{-.05in}
  \subcaption{QED}\label{subfig:QED}
\end{subfigure}
\vspace{-.05in}
\caption{\textbf{Experiments with fine-tuned generalist LLMs \citep[DrugAssist;][]{ye2023drugassist}.} Our method consistently improves the performance for generating diverse and high-quality molecules.}\label{fig:drugassist}
\vspace{-.1in}
\end{figure*}
\definecolor{whitegray}{RGB}{243, 243, 243} 

\begin{table*}[t]
\caption{\textbf{Experiments with the large number of samples.} {The base LLM is BioT5$^+$.} Our method discovers more diverse molecules with respect to the (1) the number of generations and (2) time costs.}\label{tab:time}
\vspace{-.05in}
\centering
{
\scalebox{0.92}{\begin{tabular}{lcccccc}
\toprule
Method$_{\text{num. of generations}}$ & ${\text{BS}_{300}}$ & ${\text{BS}_{400}}$ & ${\text{BS}_{500}}$ & ${\text{Ours}_{100}}$ & ${\text{Ours}_{150}}$ & ${\text{Ours}_{200}}$\\
\midrule
NCircles$_{h=0.85}$ & {13.971} & {14.553} & {15.821} & {{14.472}} & {{17.215}} & {{19.134}}  \\
Time (sec.) & {323} & {452} & {585}  & {75} & {107} & {146} \\
\bottomrule
\end{tabular}
}}
\end{table*}

\subsection{Comparison with Existing LLMs}

Additionally, we compare our fine-tuned BioT5$^{+}$ in \Cref{subsec:chebi-20} with baselines that apply existing decoding schemes to other LLMs, extending the experiment in \Cref{subsec:chebi-20}. The purpose of this experiment is to highlight that other existing LLMs have limitations in diverse molecular generation when relying on existing decoding schemes for diverse sequence generation, compared to our fine-tuned model. The overall metrics and tasks are the same as settings in \Cref{subsec:chebi-20}.

\noindent \textbf{Baselines.} For the comparison on L+M-24 dataset, we additionally consider LLMs trained on L+M-24 training split: MolT5 \citep{edwards-etal-2024-l,edwards2022molt5} and Meditron \citep{edwards-etal-2024-l,chen2023meditron}. For the comparison on ChEBI-20 dataset, we additionally consider LLMs trained on ChEBI-20 training split: MolT5 \citep{edwards2022molt5}, Text+Chem T5 \citep{christofidellis2023unifying}, and LlaSMol \citep{yu2024llasmol}. For each baseline, we report the best results (highest NCircles) obtained using one of existing decoding schemes, i.e., random sampling, BS, diverse BS, and constrastive BS. We describe detailed settings in \Cref{appx:decoding}.


\noindent \textbf{Results.} We present the results in \Cref{tab:sota}. One can see that most baselines yield a low NCircles with respect to the number of accepted and unique molecules, indicating a lack of diversity, while our method yields a relatively high NCircles.

\begin{table*}[t]
\caption{\textbf{Comparison with variants of implementations.} Our multi-stage reinforcement learning shows superior performance compared to the single-stage reinforcement learning.}\label{tab:filtering}
\vspace{-.05in}
\centering
{
\scalebox{0.92}{\begin{tabular}{lccccc}
\toprule
Method & NCircles$_{h=0.85}$ & NCircles$_{h=0.75}$ & Accepted \& Unique  & IntDiv. \\
\midrule
Div-SFT+RL$_{\text{single}}$  & {5.121} & {3.846} & {8.137} & {0.160} \\
Div-SFT+RL (ours) & {11.301} & {8.966} & {16.271} & {0.246} \\
\bottomrule
\end{tabular}
}}
\vspace{-.1in}
\end{table*}

\subsection{Fine-tuning Generalist LLMs} \label{subsec:genral_tune}

We further validate whether our fine-tuning method improves the generalist LLMs. Here, as a base LLM for implementing our method, we consider DrugAssist~\citep{ye2023drugassist} which is based on the Llama-7B \citep{touvron2023llamaopenefficientfoundation}. As baselines, we apply existing decoding schemes to DrugAssist.

\noindent \textbf{Tasks.} We consider molecular descriptions about quantitative molecular properties: hydrogen bond (HB) donors, HB acceptors, Bertz complexity \citep{bertz1981first}, and quantitative estimate of drug-likeness (QED) \citep{bickerton2012quantifying}. Note that these properties can be evaluated using external tools like RDKit \citep{landrum2025rdkit}. 

\noindent \textbf{Implementations.} We consider fine-tuning the base LLM on prompts about three properties: HB donors,  HB acceptors, and Bertz complexity. The prompts about QED are used to assess generalization to unseen properties. The overall implementations follow \Cref{subsec:chebi-20}, but the description-matching reward $r_{\text{match}}(m_k)$ is designed to yield a positive value when the given property is exactly satisfied. See \Cref{appx:implementation} for the detailed settings.

\noindent \textbf{Results.} We present the results in \Cref{fig:drugassist}. One can see that our approach improves the performance in generating diverse molecules exactly satisfying the given molecular descriptions. Furthermore, our approach consistently demonstrates superior performance for the unseen prompt, i.e., the prompt about QED in \Cref{subfig:QED}. 

\subsection{Ablation Studies} \label{subsec:abla}


\noindent \textbf{Large number of samples vs. performance.} We also analyze how well our method discovers diverse molecules with respect to the number of generations. Here, we extend beyond the settings in \Cref{subsec:chebi-20}. We consider our method to generate $100$, $150$, and $200$ molecules with a single NVIDIA A100 SXM4 40GB GPU. We also consider BS to generate a larger number of molecules with beam sizes of $300$, $400$, and $500$, reaching our maximum computational budget (four NVIDIA A100 SXM4 40GB GPUs). This experiment uses $250$ molecular descriptions in the ChEBI-20 test split. We present the results in \Cref{tab:time}. One can see that our method exhibits further performance improvements compared to the baseline, even though the baseline generates a larger number of molecules using our maximum computational budget.


\noindent \textbf{Time costs vs. performance.} In addition, we also analyze how well our method discovers diverse molecules with respect to the time costs. In \Cref{tab:time}, we present the time costs for each method. One can see that our method discovers more diverse molecules with respect to the time costs.


\noindent \textbf{Single-stage vs. multi-stage RL.} As mentioned in \Cref{subsec:RL}, we consider a multi-stage setting for generating multiple molecules. However, one may also consider a single-stage setting, where the return of a generated sequence is defined as the sum of the rewards from multiple generated molecules. In \Cref{tab:filtering}, we compare both approaches. One can see that the multi-stage setting significantly outperforms the single-stage setting. We hypothesize that this result stems from credit assignment issues in the single-stage setting. Namely, the single-stage setting lacks signals to capture molecule-wise impacts on diversity among a large set of molecules and fails to promote the generation of molecules responsible for increasing diversity.

\definecolor{whitegray}{RGB}{243, 243, 243} 

\begin{table}[h]
\vspace{-.05in}
\caption{\textbf{Coverage of the target molecules in L+M-24 dataset.} {The base LLM is BioT5$^+$.}}\label{tab:lpm}
\vspace{-.05in}
\centering
{
\scalebox{0.92}{\begin{tabular}{lcccc}
\toprule
Method & Target Coverage  \\
\midrule
Random & {0.137} \\
BS & {0.420} \\
Contrastive BS & {0.420} \\
Diverse BS & {0.343} \\
\midrule
\textbf{Ours} & {\textbf{0.558}} \\
\bottomrule
\end{tabular}}
}
\vspace{-.05in}
\end{table}

\noindent \textbf{Coverage of the target molecules.} We further validate how our method benefits capturing the wide range of target molecular space. Here, we compute how many target molecules in the L+M-24 dataset (an average of $17$ molecules per description) can be captured within the generation space. The target molecule is considered captured if it is significantly similar to one of the generated molecules (Dice similarity $>0.95$). In \Cref{tab:lpm}, we present the average ratio of captured target molecules. One can see that our method yields the highest score. 

\section{Conclusion}
\label{sec:conclusion}

In this paper, we identify the limitations of large language models (LLMs) for generating diverse molecules. In response, we present a new fine-tuning approach to adapt existing LLMs to generate diverse molecules. Experiments show that our approach enables LLMs to better discover diverse molecules compared to the existing approaches. This success highlights the potential of our method to advance LLM-driven drug discovery.

\section{Limitations} 

Our method may require autoregressively generating a very long sequence to induce a huge set of molecules. In response, an interesting avenue for future work is to reduce the space and time complexity in generating the sequence of molecules, for example, introducing continuous tokens \citep{hao2025training} that encode the set of previously generated molecules \citep{NIPS2017_f22e4747}. Next, we validated whether the LLM-generated molecules satisfy the given molecular descriptions using computational approaches. However, this does not imply that the generated molecules will necessarily satisfy the given molecular descriptions in tricky real-world scenarios. To ensure the reliability of our method, the generated molecules should be validated through the real-world experiments. Additionally, although the molecules generated by each method are valid in computational terms, they may not be synthesizable in practice. Lastly, due to the limited computational budgets, we conducted experiments only on models with up to 7B parameters. The generalizability of our method to larger models (e.g., 70B) remains unexplored and is left for future work.

\section{Ethical Considerations} 

Our fine-tuning method enables LLMs to generate diverse molecules from textual descriptions of molecular properties. However, these advancements also introduce potential risks, such as the generation of harmful drugs and the misuse of synthesized molecules.




\bibliography{iclr2025_conference}
\newpage

\appendix
\onecolumn

\section{{Additional Related Works}} \label{appx:addition_related}

\subsection{{Reinforcement Learning (RL) for Diverse Molecular Generation}}

{Existing literature has studied RL-based methods to generate molecules with desired properties while enhancing their diversity. First, \citet{blaschke2020memory,pereira2021diversity} introduced memory-assisted RL, which penalizes the reward of a molecule when it is highly similar to the molecules stored in the memory unit. \citet{he2024evaluation} also incorporated RL with a diversity penalty in transformer-based architectures for molecular generation. In addition, \citet{hu2024novo} leveraged multiple GPT-based agents trained with RL to encourage these agents to explore diverse directions for discovering diverse molecules. Their algorithms are designed to consider a fixed target property. In contrast, our work fine-tunes LLMs to generate diverse molecules given a prompt that is flexible to describe various target properties.}

\subsection{{Molecular Similarity Measures}} \label{appx:diversity}

In this section, we explain the structural similarity between two molecules. These measures are used to define the reward of reinforcement learning (\Cref{appx:implementation}) and diversity metrics (\Cref{appx:ncircles}). 

\noindent \textbf{Molecular structural features.} First, molecular structural features are expressed with Morgan fingerprint \citep{rogers2010extended}, which is a vector that characterizes the presence of specific atoms, bonds, or substructures in a given molecule. Note that we get this from RDkit package \citep{rdkit}. We denote this with $f(m)$, which maps the molecule $m$ to its Morgan fingerprint. Next, we compute the molecular structural similarity based on the molecular fingerprints. 

\noindent \textbf{Tanimoto similarity.} The overall similarity between two molecules is typically evaluated as follows:
\begin{equation*}
T(m_i,m_j)=\frac{|f(m_i)\cap f(m_i)|}{|f(m_i)\cup f(m_j)|},
\end{equation*}
where $f(m_i)$ maps the molecule $m_i$ to its Morgan fingerprint. This similarity $T(m_i,m_j)$ is referred to as Tanimoto similarity \citep{bajusz2015tanimoto}, evaluating overall structural similarity.

\noindent {\textbf{Dice similarity.} In this paper, we also consider Dice similarity \citep{bajusz2015tanimoto} which is sensitive to the degree of shared structural features between two molecules. This is defined as follows:}
{\begin{equation*}
D(m_i,m_j)=\frac{2|f(m_i)\cap f(m_i)|}{|f(m_i)|+ |f(m_j)|},
\end{equation*}}
{where $\langle \cdot,\cdot \rangle$ denotes a dot product between two vectors.}


\subsection{{Molecular Diversity Metrics}}\label{appx:ncircles}

{In this section, we provide a detailed explanation of the diversity metrics for evaluating the given set of molecules. Specifically, we explain two diversity metrics: the number of circles \citep{xie2023how} and the internal diversity \citep{moses}.}

\noindent {\textbf{The number of circles \citep[NCircles.;][]{xie2023how}.} To evaluate the diversity of a given set of molecules $\mathcal{M}$, this computes the size of the largest subset of molecules in which no two molecules are similar to each other. This metric is defined with a Tanimoto similarity $T(\cdot,\cdot)$ and a similarity threshold $h$ as follows:}
\begin{equation}{
\text{NCircles}_{h}=\text{max}_{\mathcal{C}\subseteq\mathcal{M}}\;|\mathcal{C}|\quad \text{s.t. }T(x,y)<h, \forall x \neq y \in \mathcal{C},}
\end{equation}
{where $\mathcal{C}$ is a subset of molecules. Every pair of molecules in $\mathcal{C}$ should have a similarity lower than $h$. The high NCircles value implies that the given set of molecules $\mathcal{M}$ is diverse and covers a wide range of molecular space \citep{xie2023how}. Recent work \citep{renz2024diverse} haven shown that this metric is relatively exact to measure the molecular diversity, compared to the other metrics, e.g., internal diversity.

\noindent {\textbf{Internal diversity \citep[IntDiv.;][]{moses}.} Given a set of molecules $\mathcal{M}$, this metric measures the average of pair-wise Tanimoto similarities:}
\begin{equation}{
\text{Intdiv.}=\frac{1}{|\mathcal{M}| \cdot (|\mathcal{M}| - 1)} \sum_{i=1}^{|\mathcal{M}|} \sum_{j=i+1}^{|\mathcal{M}|} (1-T(m_i, m_j)),}
\end{equation}
{where $m_i$ is $i$-th molecule in the given set of molecules $\mathcal{M}$.}

\newpage

\section{Detailed Implementations and Training} \label{appx:implementation}

\subsection{Supervised Fine-tuning}

\textbf{Dataset collection.} To fine-tune BioT5$^{+}$ (\Cref{subsec:chebi-20}), we collect $T=100$ molecules using contrastive beam search for each training molecular description in ChEBI-20 training split. In the case of molecular descriptions from the L+M-24 training split, we use only the provided set of target molecules for each description, without additional data collection. To fine-tune DrugAssist (\Cref{subsec:genral_tune}), we collect $T=300$ molecules using contrastive beam search for each training prompt. Note that the collected molecules were filtered to remove invalid string representations, duplicated molecules, and unaccepted molecules. The invalid string representations are evaluated with RDKit package \citep{rdkit}. Additionally, the collected molecules are concatenated into a single sequence $m_1||\cdots||m_K$. 

\noindent \textbf{Supervised learning.} We consider four NVIDIA A100 SXM4 40GB GPUs for supervised fine-tuning. 
\begin{itemize}[topsep=-1.0pt,itemsep=1.0pt,leftmargin=3.5mm]
\item For the supervised fine-tuning of BioT5$^{+}$ (\Cref{subsec:chebi-20}), we consider $80$ epochs, $8$-batch size, $5\textrm{e}-4$ learning rate, $0.05$ warm-up ratio, and apply a cosine learning scheduler. The maximum sequence length in supervised training is limited to $2560$ due to memory limitations.
\item For the supervised fine-tuning of DrugAssist (\Cref{subsec:genral_tune}), we consider $80$ epochs, $4$-batch size, $3\textrm{e}-5$ learning rate, $0.05$ warm-up ratio, and apply a cosine learning scheduler. The maximum sequence length in supervised training is limited to $1024$ due to memory limitations. We also apply LoRA \citep{hu2022lora}, where the rank and alpha are $64$ and $128$, respectively.
\end{itemize}

\subsection{Reinforcement Learning}

\noindent \textbf{Reward design.} In experiments on L+M-24 and ChEBI-20 datasets (\Cref{subsec:chebi-20}), we define the description-matching reward using the Dice similarity as follows: 
\begin{equation}\label{eq:match}
r_{\text{match}}(m_k,p_{\text{desc}})=\max_{m_{\text{target}\in \mathcal{M}_{\mathcal{D}}(p_{\text{desc}})}} D(m_{\text{target}}, m_k)^{\alpha}, 
\end{equation} 
where $\mathcal{M}_{\mathcal{D}}(p_{\text{desc}})$ is a set of target molecules provided in the datasets for each molecular description $p_{\text{desc}}$.\footnote{As described in \Cref{subsec:chebi-20}, we assume that the molecule satisfies the description if it shares a certain degree of structures with one of the target molecules provided in the dataset.} As described in \Cref{appx:diversity}, $D(m_i,m_j)$ is Dice similarity captuing the degree of shared structural features between two molecules. Note that $\alpha$ is a hyper-parameter. In experiments with DrugAssist (\Cref{subsec:genral_tune}), the description-matching reward yields $1$ if the molecule satisfies the quantitative properties described in $p_{\text{desc}}$ (evaluated using RDkit \citep{rdkit}) and $0$ otherwise.

Next, the diversity reward, $r_{\text{div}}$, is defined to consider molecular structural diversity between the new molecule $m_k$ and the previously generated molecules $\{m_i\}^{k-1}_{i=1}$ within a sequence: 
\begin{equation} \label{eq:div}
r_{\text{div}}(m_k,\{m_i\}^{k-1}_{i=1})=1-\text{max}_{m \in \{m_i\}^{k-1}_{i=1}}T(m_k,m)^{\beta}, 
\end{equation} 
where $T(m_i,m_j)$ is Tanimoto similarity (\Cref{appx:diversity}) between $m_i$ and $m_j$ and $\beta$ is a hyper-parameter.

\noindent \textbf{Policy optimization.} We consider four NVIDIA A100 SXM4 40GB for reinforcement learning implemented with proximal policy optimization. 

\begin{itemize}[topsep=-1.0pt,itemsep=1.0pt,leftmargin=3.5mm]
\item For the reinforcement learning of BioT5$^{+}$ (\Cref{subsec:chebi-20}), we consider $200$ PPO iterations, $8$ mini-batch size, $128$ batch size, and $5\textrm{e}-5$ learning rate. We also consider $0.01$ KL penalty. Note that $\alpha$ and $\beta$ in \Cref{eq:match,eq:div} are $0.5$ and $1.0$, respectively. We increate $\beta$ to $2.0$ during training on ChEBI-20 dataset. The reward signal is amplified by multiplying by a value of $8.0$. The maximum sequence length in reinforcement learning is limited to $2560$ due to memory limitations. Here, we also apply LoRA \citep{hu2022lora} where the rank and alpha are $16$ and $32$, respectively. 
\item For the reinforcement learning of DrugAssist (\Cref{subsec:genral_tune}), we consider $200$ PPO iterations, $4$ mini-batch size, $64$ batch size, and $3\textrm{e}-6$ learning rate. We also consider $0.1$ KL penalty. Note that $\beta$ in \Cref{eq:div} is $2.0$. The reward signal is amplified by multiplying by a value of $8.0$. The maximum sequence length in reinforcement learning is limited to $1024$ due to memory limitations. We also apply LoRA \citep{hu2022lora} where the rank and alpha are $64$ and $128$, respectively. 
\end{itemize}

\newpage

\section{Dataset Details} \label{appx:stat}

\begin{wrapfigure}{r}{0.34\textwidth}
\centering
\vspace{-.2in}
\captionof{table}{\textbf{Data statistics}. The number of molecular descriptions.}\label{tab:stats}
\vspace{-.05in}
\scalebox{0.95}{\begin{tabular}{lcc}
\toprule
Dataset & \multicolumn{1}{c}{$\sharp$ of training} & \multicolumn{1}{c}{$\sharp$ of test} \\
\midrule
L+M-24 & {160,492} & {21,839} \\
ChEBI-20 & {26,408} & {3,301} \\
\bottomrule
\end{tabular}}
\end{wrapfigure}

In this section, we describe detailed data statistics of ChEBI-20 \citep{edwards-etal-2021-text2mol} and L+M-24 datasets \citep{edwards-etal-2024-l}. Both datasets serve to validate description-guided molecular generation capability. Note that both datasets consider a molecular description expressed in English and a molecule represented by SMILES \citep{weininger1988smiles}. In \Cref{tab:stats}, we provide the number of molecular descriptions.

To be specific, L+M-24 dataset provides an average of $70$ and $17$ target molecules for each description in training and test splits, respectively. Note that the original L+M-24 dataset provides a single target molecule for each description. However, L+M-24 dataset involves identical molecular properties for different molecules. Moreover, some molecular descriptions, e.g., an inhibitor of both BCL2 and BTK proteins, include the molecular properties of other descriptions, e.g., an inhibitor of BCL2 protein. Thus, we associate multiple molecules with a single molecular description. Next, ChEBI-20 dataset provides a single example of target molecule for each description. 

In addition, since the original BioT5$^{+}$ is not pre-trained on L+M-24 dataset, we further train it on the original L+M-24 training split. Note that we use the first $1,000$ molecular descriptions in the test split of the L+M-24 dataset for evaluation, as evaluating the entire test split with $50$ generated molecules per description takes too much time (five to six days).

\newpage

\section{Experiments Setup}

Our experiments consider various molecular generative LLMs: MolT5 \citep{edwards2022molt5}, BioT5$^{+}$ \citep{pei2024biot5plus}, Text+Chem T5 \citep{christofidellis2023unifying}, LlasMol \citep{yu2024llasmol}, DrugAssist \citep{ye2023drugassist}, and Meditron \citep{chen2023meditron}.\footnote{MolT5, BioT5$^{+}$, Text+Chem T5, LlasMol, and DrugAssist are licensed under the MIT License.}\footnote{Meditron is licensed under the Apache 2.0 License.} They consider a molecular description expressed in English. They also consider a molecule represented as SMILES \citep{weininger1988smiles}, except for BioT5$^{+}$, which uses SELFIES \citep{krenn2020self}. T5 models have 220M parameters, and others have 7B parameters. We use four NVIDIA A100 SXM4 40GB GPUs for the experiments. We consider a single run.

\noindent \textbf{Comparison with decoding schemes.}\label{appx:decoding}\label{appx:sota} In this experiment, we first consider random sampling with different temperatures $\{0.7, 1.0, 1.5\}$. For the other decoding schemes, we consider conventional configurations: nucleus sampling with top-p $0.8$, beam search, diverse beam search with a diversity penalty of $0.5$, and contrastive beam search with a penalty alpha of $0.5$. We apply greedy decoding for our approach. The prompts for BioT5$^{+}$ are described in \Cref{tab:prompts_biot5}.

\begin{table}[h]
\caption{\textbf{Prompts for BioT5$^{+}$ \citep{pei2024biot5plus}.}} \label{tab:prompts_biot5}
\centering
\scalebox{0.88}{
\begin{tabular}{l m{0.7\textwidth}}
\toprule
Prompt  & Contents  \\
\midrule
$p_{\text{desc}}$ & ``Definition: You are given a molecule description in English. Your job is to generate the molecule SELFIES that fits the description. \par\par Now complete the following example - \par Input: <molecular description> \par Output: ''\\
\midrule
$p_{\text{desc+div}}$ (fine-tuning) & ``Definition: You are given a molecule description in English. Your job is to generate the molecule SELFIES that fits the description. \par\par Now provide a set of molecules - \par Input: <molecular description> \par Output: ''\\
\bottomrule
\end{tabular}}
\end{table}

\noindent \textbf{Comparison with existing LLMs.} For all existing LLMs, we apply random sampling, beam search, diverse beam search with a diversity penalty of $0.5$, and contrastive beam search with a penalty alpha of $0.5$. The prompts for existing LLMs, which are molecular generative LLMs based on a given molecular description, are described in \Cref{tab:galatica}.


\begin{table}[h]
\caption{\textbf{Prompts for various existing LLMs}} \label{tab:galatica}
\centering
\scalebox{0.88}{
\begin{tabular}{l m{0.77\textwidth}}
\toprule
Method  & $p_{\text{desc}}$  \\
\midrule
MolT5 &  ``<molecular description>'' \\
\midrule
Meditron & ``Below is an instruction that describes a task, paired with an input that provides further context. Write a response that appropriately completes the request. \par \#\#\# Instruction: \par You are a researcher. You can come up molecule smile strings based on your existing knowledge. \par Molecule smile strings are given against the following input. You should be as detailed as possible. \par\par Input: \par <molecular description>  \par In that caption, could you generate a molecule smile string? \par \#\#\# Response: '' \\
\midrule
Text+Chem T5 & ``<molecular description>'' \\
\midrule
LlasMol & ``Give me a molecule that satisfies the conditions outlined in the description: <molecular description>''\\
\bottomrule
\end{tabular}}
\end{table}

\newpage

\begin{table}[H]
\caption{\textbf{Prompts for DrugAssist \citep{ye2023drugassist}.}} \label{tab:drugassist}
\centering
\scalebox{0.88}{
\begin{tabular}{l m{0.7\textwidth}}
\toprule
Prompt  & Contents  \\
\midrule
$p_{\text{desc}}$ & Hydrogen bond donors and acceptors: ``Can you generate a molecule with <value> <property>? Print it in SMILES format.'' \par QED and Bertz complexity: ``Can you generate a molecule with <property> below <value1> but at least <value2>? Print it in SMILES format.''\\
\midrule
$p_{\text{desc+div}}$ (fine-tuning) & Hydrogen bond donors and acceptors: ``Can you generate a set of molecules that have <value> <property>? Print each of them in SMILES format.'' \par QED and Bertz complexity: ``Can you generate a set of molecules that have <property> below <value1> but at least <value2>? Print each of them in SMILES format.''\\
\bottomrule
\end{tabular}}
\end{table}

\begin{table}[H]
\caption{\textbf{Prompts for DrugAssist (without fine-tuning, \Cref{fig:example}).}} \label{tab:design}
\centering
\scalebox{0.88}{
\begin{tabular}{m{0.97\textwidth}}
\toprule
$p_{\text{desc+div}}$  \\
\midrule
``Can you generate a set of molecules? Each molecule has <value> <property>. Print each of them in SMILES format.'' \\
\midrule
``Can you generate a diverse set of molecules? Each molecule has <value> <property>. Print each of them in SMILES format.'' \\
\midrule
``Can you generate a structurally diverse set of molecules? Each molecule has <value> <property>. Print each of them in SMILES format.'' \\
\midrule
``Can you generate a set of molecules with <value> <property>? Print each of them in SMILES format.'' \\
\midrule
``Can you generate a diverse set of molecules with <value> <property>? Print each of them in SMILES format.'' \\
\midrule
``Can you generate a structurally diverse set of molecules with <value> <property>? Print each of them in SMILES format.'' \\
\bottomrule
\end{tabular}}
\end{table}

\noindent \textbf{Fine-tuning generalist LLMs.} We synthesize $600$ pairs of training prompts and corresponding sets of molecules. The prompts specify hydrogen bond donors and acceptors ranging from one to four, and a Bertz complexity ranging from $0$ to $300$. The sets of molecules are collected by applying beam search on DrugAssist and then perturbed by shuffling their order. We use the prompts described in \Cref{tab:drugassist}. The system prompt follows the default settings from \citep{ye2023drugassist}. For evaluation, we use four prompts specifying three hydrogen bond donors, three hydrogen bond acceptors, a Bertz complexity between $100$ and $200$, and a QED value between $0.4$ and $0.6$. Additionally, as shown in \Cref{fig:example}, we try to generate diverse molecules by designing prompts (\Cref{tab:design}). However, these prompts show lower performance compared to applying beam search.

\newpage

\section{Additional Results} \label{appx:qualit}

In this section, we present qualitative results by evaluating whether the generated molecules satisfy the given molecular description using an external tool. Here, we consider the generation of a set of molecules from: ``The molecule is a egfr
inhibitor and modulate, belonging to the cancer treatment class of molecules, and is treatment of disorder'' obtained from L+M-24 test split. Note that the molecule with sufficient binding affinity to EGFR can satisfy this description. Thus, we evaluate the generated molecule using a docking tool \citep{garcia2022dockstring}. We accept the generated molecule if it yields a Vina score \citep{trott2010autodock} below $-7.0$ when docked with an EGFR protein, which is the threshold for a potential inhibitor of the protein. We present the results in \Cref{tab:BS_example}. One can see that BS generates structurally similar molecules while our method generates structurally diverse molecules.

\begin{table}[t]
\caption{\textbf{Visualization of the generated molecules.} We generate $50$ molecules from: ``The molecule is a egfr inhibitor and modulate, belonging to the cancer treatment class of molecules, and is treatment of disorder''. The \textcolor{blue}{blue line} indicates the molecule that follows the given description, as evaluated using the external tool \citep[docking tool;][]{trott2010autodock}. We also evaluate \sethlcolor{OrangeRed}\hl{structurally similar molecules} (Tanimoto similarity $>0.7$).}
\label{tab:BS_example}
\vspace{-.05in}
\centering
\scalebox{0.88}{
\begin{tabular}{c}
\toprule
$50$ molecules generated from BS \\
\midrule
\includegraphics[width=0.97\linewidth]{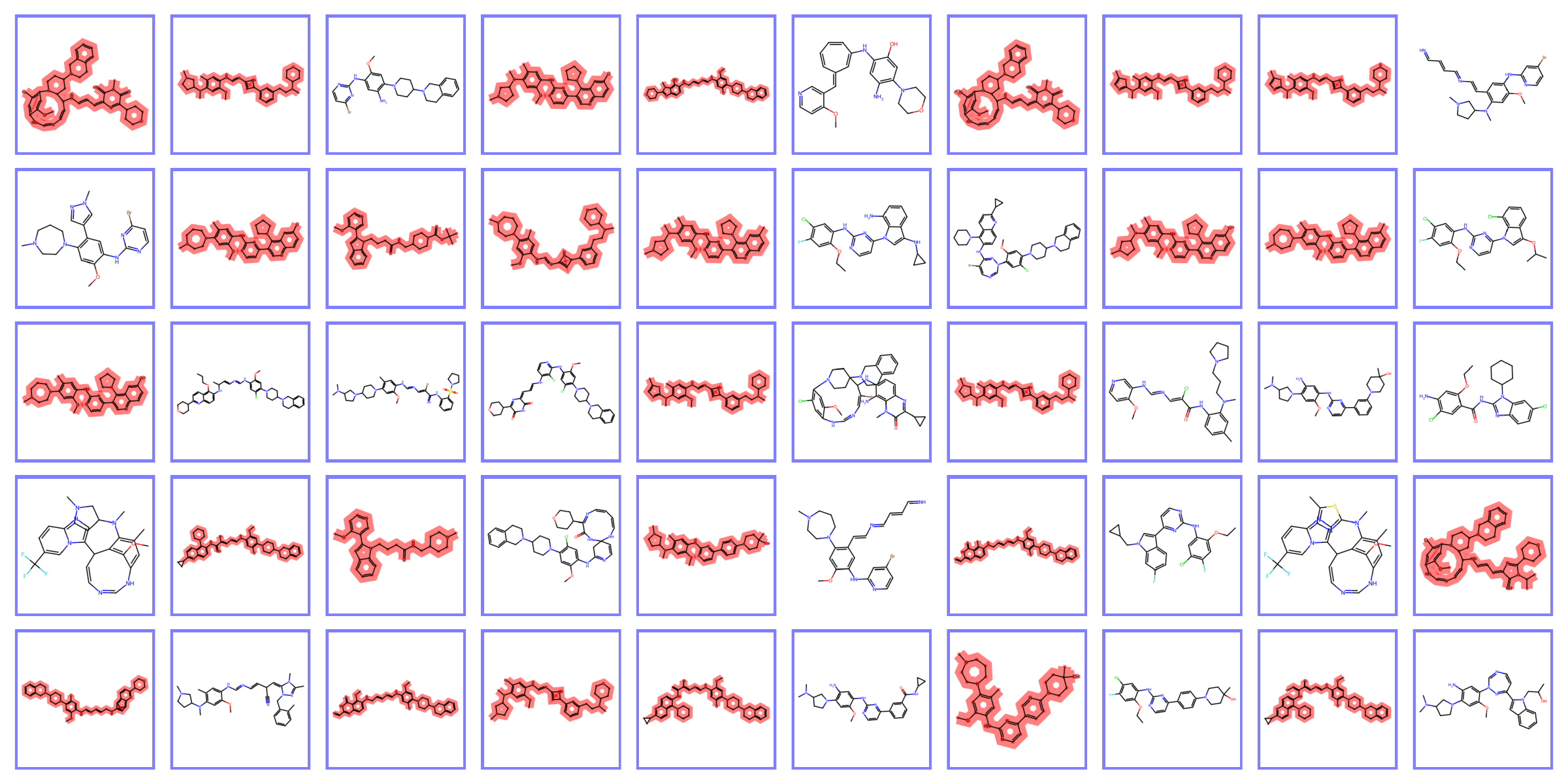} \vspace{-.03in}\\
\midrule
$50$ molecules generated from ours \\
\midrule
\includegraphics[width=0.97\linewidth]{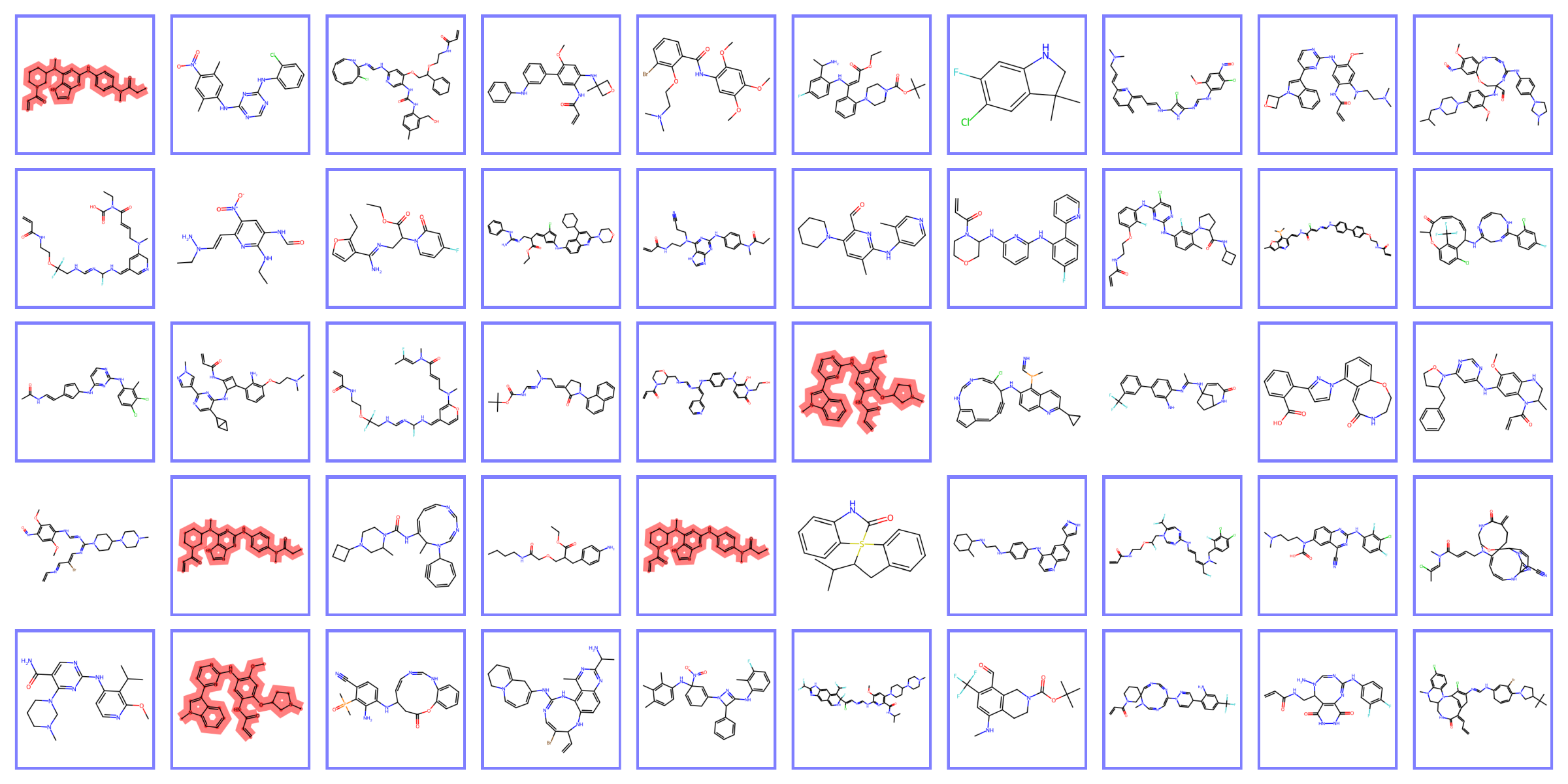}\vspace{-.03in} \\
\bottomrule
\end{tabular}}
\end{table}

\newpage

\section{Use of AI Assistants}

We used AI-based writing assistants to improve the sentence structure and grammar. These tools were used only for editorial improvements. The technical content, methodology, and experimental results were entirely authored by the researchers.

\end{document}